\documentclass[conference]{IEEEtran}
\usepackage{graphicx}
\usepackage{epstopdf}
\usepackage{subcaption}
\usepackage{color}
\usepackage{cite}
\usepackage{balance}

\begin{document}
\title{Efficient and accurate monitoring of the depth information in a Wireless Multimedia Sensor Network based surveillance}

\author{\IEEEauthorblockN{Anthony Tannoury and Rony Darazi}
\IEEEauthorblockA{TICKET Lab. 	\\
Antonine University\\
	Hadat-Baabda, Lebanon.\\
Email: anthony.tannoury@ua.edu.lb}

\and
\IEEEauthorblockN{Christophe Guyeux and Abdallah Makhoul}
\IEEEauthorblockA{Femto-ST Institute, UMR 6174 CNRS\\
Universit\'{e} de Bourgogne Franche-Comt\'{e}\\
France\\
}}
\maketitle

\begin{abstract}	
Wireless Multimedia Sensor Network (WMSN) is
a promising technology capturing rich multimedia data like audio and video, which can be useful to 
monitor an environment under surveillance. However, many scenarios in real time monitoring requires 3D depth
information. In this research work, we propose to use
the disparity map that is computed from two or multiple images,
in order to monitor the depth information in an object or event under surveillance using WMSN. Our system is based on distributed wireless sensors allowing us to notably reduce the computational time needed for 3D depth reconstruction, thus permitting the success of real time solutions. Each pair of sensors will capture images for a targeted place/object and will operate a Stereo Matching in order to create a Disparity Map. Disparity maps will give us the ability to decrease traffic on the bandwidth, because they are of low size. This will increase WMSN lifetime. Any event can be detected after computing the depth value for the target object in the scene, and also 3D scene reconstruction can be achieved with a disparity map and some reference(s) image(s) taken by the node(s).
\end{abstract}

\IEEEpeerreviewmaketitle

\section{Introduction}
Monitoring and surveillance attract many researchers, specially with the development of the Internet of Things (IoTs). Within this discipline, Wireless Senor Networks (WSNs) are able to sense scalar data like temperature, light, and so on~\cite{Sharif09}, while 
Wireless Multimedia Sensor Networks add the access to audio, images, or video data.  They are built with low cost~\cite{bbdgm16} Complementary Metal Oxide Semiconductor (CMOS) cameras and have a big range of applications in different fields like health, military, environmental, etc. 
The main challenges in WMSNs is to capture and process needed information with low energy consumption, fast computation time, and high Quality of Service (QoS).

The depth of an object or a monitored character gives us an essential clue for tracking or triggering an event. This depth will create a 3d representation  of any object and it cannot be supplied by conventional gathered data.
Most existing 3D reconstruction methods~\cite{Ripolles2014} use professional, complex, and expensive sensors with a specific network topology to obtain the required results.
However, camera sensors in WMSNs have low energy resources because they are powered by limited batteries, so energy consumption is a primary constraint for WMSNs as we aim to prevent from decreasing uselessly the network lifetime. Our main objective is thus to compute 3D-depth while maintaining low power consumption, high speed, and QoS in the specific context of wireless multimedia sensor networks. 

This is why we propose the use of \textit{low-complexity} disparity maps computed by a couple of \textit{low cost} sensor nodes in a WMSN. This is done to efficiently provide, to the monitoring system, an important depth information of the object under surveillance. Indeed, having multiple images captured at low cost from different views, and applying Stereo Vision on different couples after regrouping each pair of sensors as left camera and right one, our system can calculate at low cost some disparity maps to recover the depth information.

\begin{figure}[t]
	\centering
	\includegraphics[scale=0.2]{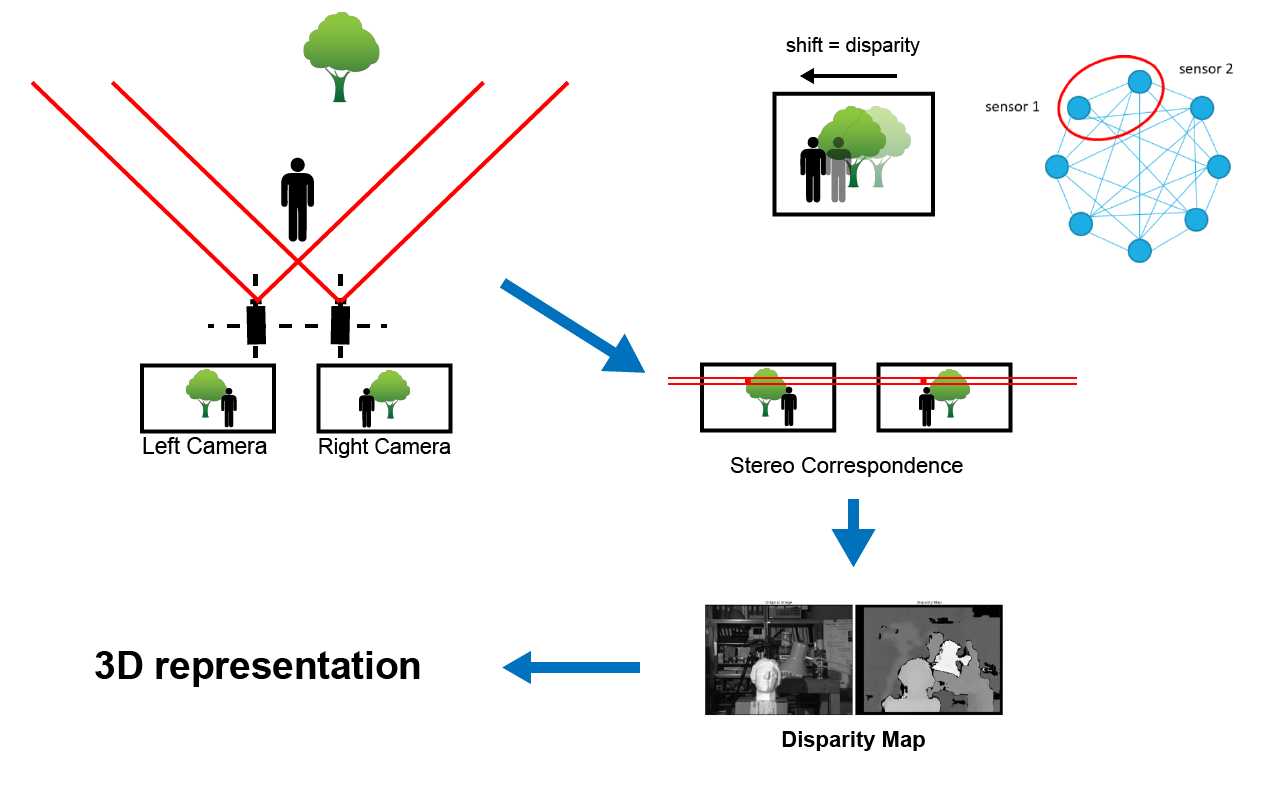}
	\caption{Phases of our system}
	\label{fig:figure1} % always label after caption 
	
\end{figure}

This system is summarized in Figure~\ref{fig:figure1}. As can be seen, data processing is made locally in each couple of sensors. This will reduce the transmission rate that has a direct and important effect on the network lifetime. Our system will thus not transmit high resolution images, but only gray-scale disparity maps, and on demand -- depending on the triggered event or changes in the scene.

The remainder of this article is organized as follow. Section~\ref{sec:WMSN} introduces the WMSN specifications, challenges, and limitations. In Section~\ref{sec:DMmethods}, two low cost disparity map extraction methods are presented, namely the Sum of Absolute Differences (SAD) and Sum of Squared Differences (SSD) ones. They are expedimented in the next section, and a complexity estimation is provided. This research article ends by a conclusion section, in which the contribution is summarized and intended future work is outlined.

\section{Wireless multimedia sensor networks}
\label{sec:WMSN}

\subsection{Composition of WMSNs}
WMSN are primarily built with wireless sensors able to record images, audios, and videos. They are powered by batteries and they have the capability of sensing, processing, and transmitting data~\cite{Almalkawi10}. Finally, the availability of low cost CMOS cameras and the development of low complexity signal processing technologies and algorithms gives to the WMSNs the ability to provide useful multimedia data at low cost. 

\subsection{Applications of WMSNs}

WMSNs can be used in various applications for divergent fields like object tracking, agricultural monitoring, e-health, and so on~\cite{Ang:13}.
For instance, it can be deployed in a scene to monitor visible and hidden objects from different angles. Ham \textit{et al.}~\cite{Ham2016} and Golparvar-Fard \textit{et al.}~\cite{Golparvar11} supervised buildings and constructions. Augmented reality can be mixed too with WMSNs to visualize real time some 3d representations using mobile software~\cite{Goldsmith08}.

\subsection{WMSN challenges}
The main constraint in WMSN is energy consumption, because sensors are powered by small batteries. So optimizing energy is very essential in our work independently from the type of application. Most of the energy is consumed by transmission. So decreasing the transmission rate and distance between nodes will increase network lifetime.
As an illustration, 64KB data processed in a wireless sensor node leads to a consumption of $0.00195\mu J$ for program execution (data processing), which has no comparison with the  $377\mu J$ of power required for data transmission (radio), as experimented in~\cite{Zhao2013}.

WMSNs need higher data rate while using high resolution images. 
Capturing images in a short time can be achieved but video streaming requires continuous capturing and delivery. In some monitoring scenarios, the system should work on real-time. Hence, powerful hardware and software techniques are a must to deliver the  QoS requested by the considered applications.

\subsection{Problematic and Solution}
Our main problematic is how to develop an efficient WMSN-based monitoring that runs for a long period of time, works in real-time, while guaranteeing a good QoS~\cite{bgmp12:ij,begh+16:ij}? In the studied context, the transferred data will be operated later for 3D scene reconstruction, at sink level. Stereo matching is done on each pair of sensors in order to deliver low size disparity maps saving depth information. Our system uses distributed WMSN, so the disparity map processing is shared recursively on all sensor nodes during the transfer to the sink, in order to decrease network load and energy consumption.

The next section outlines two low cost disparity map calculation methods that can reasonably be applied on WMSN.

\section{Disparity Map Calculation on WMSN}
\label{sec:DMmethods}
Disparity map shows the pixels difference or motion between two stereo images captured from two (left and right) sensors. Being an hot topic, new developments and techniques are introduced each year to improve the quality of the produced maps. We are however not focusing on the most up-to-date methods of high quality, as they are complex, but we intend to choose the most appropriate disparity map method in the WMSN context. In other words, we do not target a disparity map of the best quality, but the best compromise between quality of this latter and complexity to obtain it. 

Existing methods can be classified in two main categories, namely the local methods and the global ones.

In our previous work~\cite{Tannoury}, we have investigated all reputed disparity map calculation methods. And we have chosen the sole local methods as unique convenient solutions for our WMSNs context, in terms of low complexity, fast speed, and reasonable quality. It is indeed well known~\cite{yoon2006adaptive} that global methods achieve good results, but are computationally expensive. Besides, local methods, especially the Sum of Absolute Differences (SAD) and Sum of Squared Differences (SSD) ones, bring off reasonable results while being low cost.

In SAD algorithm, the absolute difference between the intensity of each pixel in the reference block and the one of the corresponding pixel in the target block is computed with the following formula:

\begin{equation} \label{eq:3}
SAD (x,y,d) = \displaystyle\sum_{(x,y)\in w}{\left| I_{l}(x,y) - I_{r}(x-d,y) \right|}
\end{equation}

SAD makes the sum of differences over $w$, where $w$ is the aggregated support window.
The SSD algorithm, for its part, can be summarized as follows, see Equation~\ref{eq:4}:

\begin{equation} \label{eq:4}
SSD (x,y,d) = \displaystyle\sum_{(x,y)\in w}{\left| I_{l}(x,y) - I_{r}(x-d,y) \right|^2}
\end{equation}

We now intend to extend the aforementioned study by a deep experimentation and complexity calculation, to choose a good trade-off between speed, computation cost, and disparity map quality.
The experiment and complexity computations, provided in the next section, ensure how Sum of Absolute Differences (SAD) is better than Sum of Squared Differences (SSD) with respects to computation time and complexity.

\section{Experiments and Complexity Computation}
\label{sec:experimentation}

\begin{figure}[t]
	
	\centering
	\includegraphics[scale=0.5]{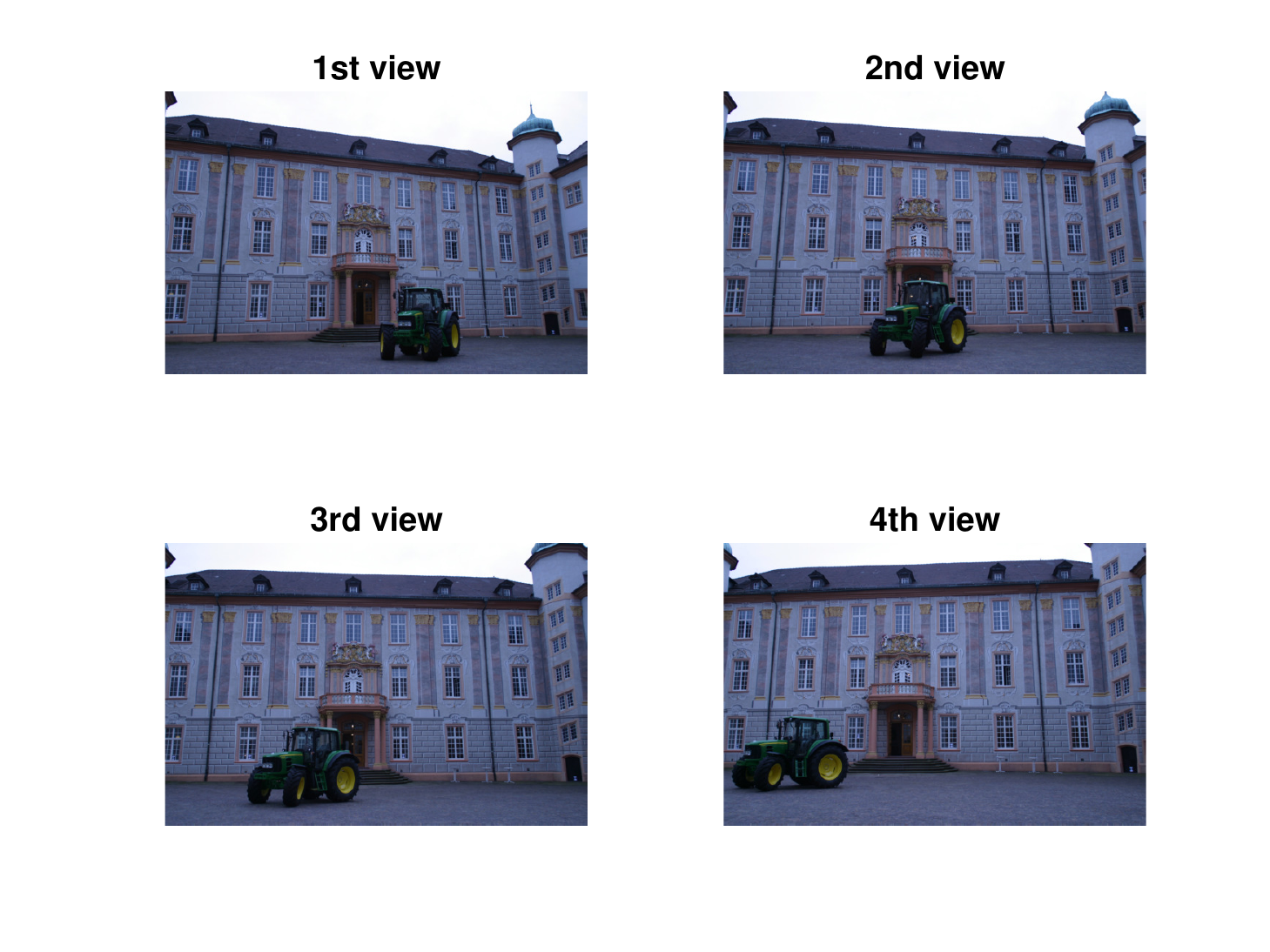}
	\caption{Different views of the scene}
	\label{fig:Figure2}% always label after caption 
	
\end{figure}

\begin{figure}[t]
	
	\includegraphics[scale=0.2]{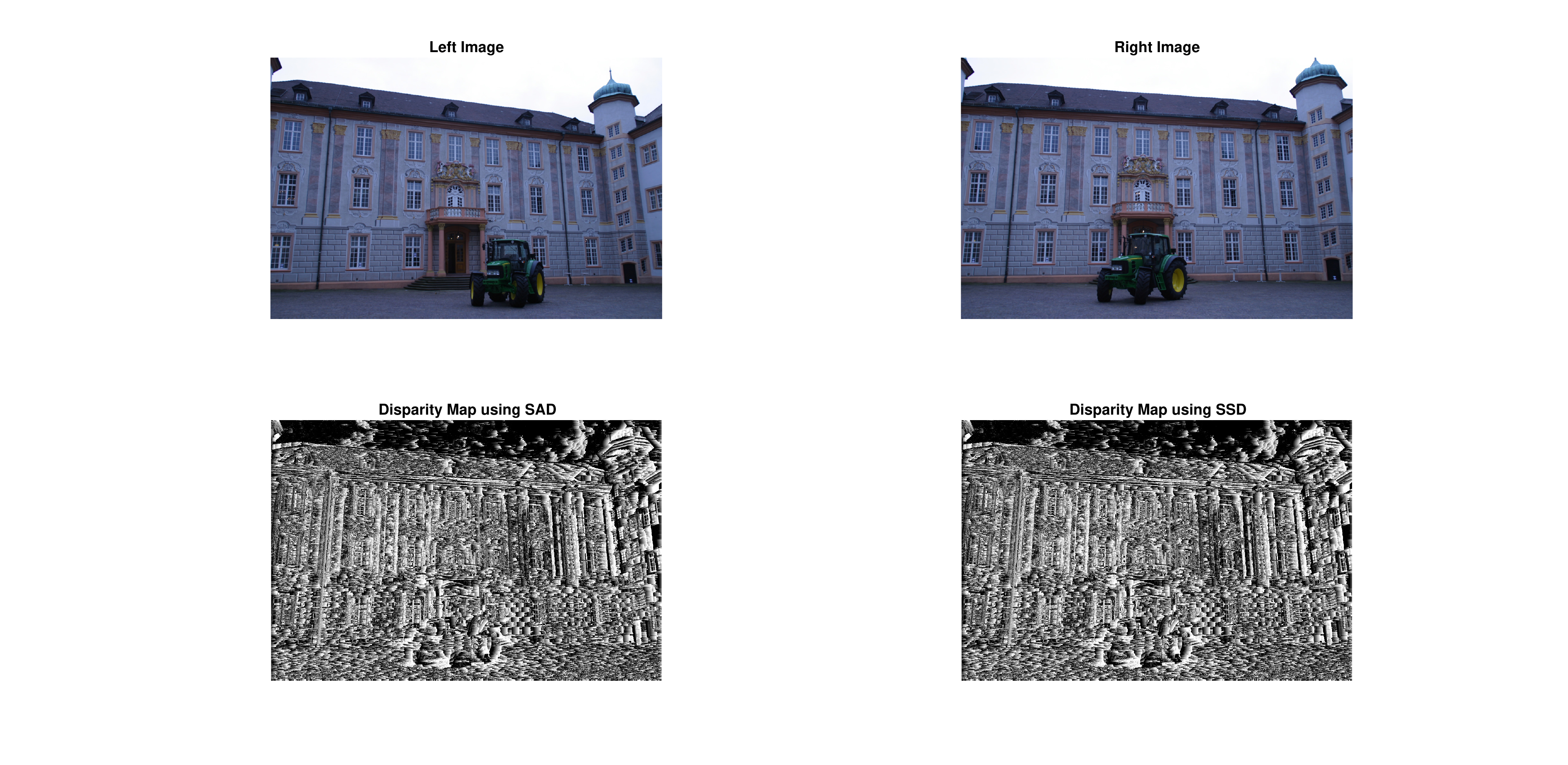}
	\caption{Disparity Map using SAD and SSD}
	\label{fig:Figure3}% always label after caption 
	
\end{figure}

\begin{figure}
	
	\begin{subfigure}{0.5\textwidth}
		\includegraphics[width=\linewidth]{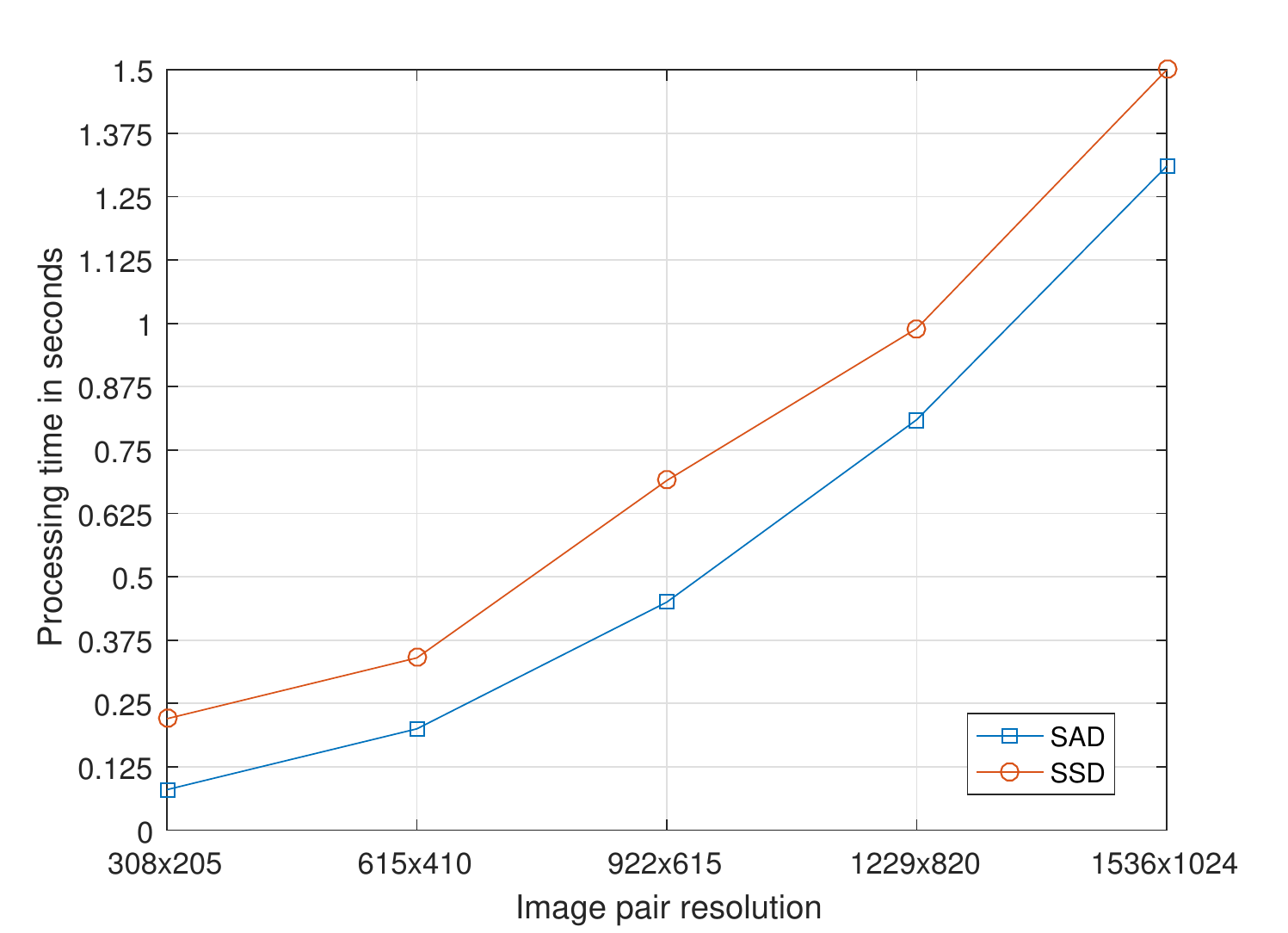}
		\caption{Processing Time with different image resolutions}
		\label{fig:sub1}
	\end{subfigure}
	
	\begin{subfigure}{0.5\textwidth}
		
		\includegraphics[width=\linewidth]{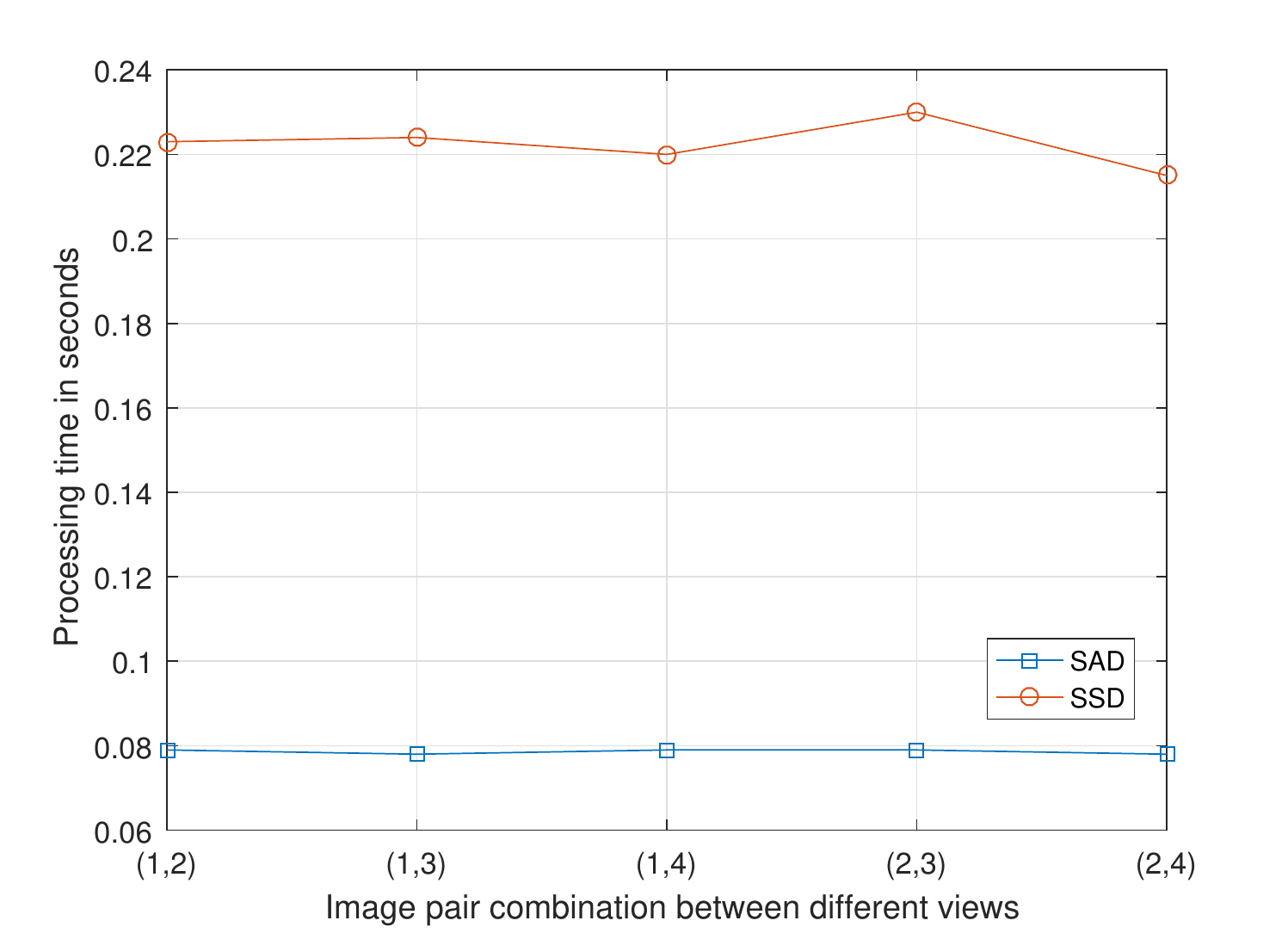}
		\caption{Processing Time with different coupling}
		\label{fig:sub2}
	\end{subfigure}
	
	\begin{subfigure}{0.5\textwidth}
		
		\includegraphics[width=\linewidth]{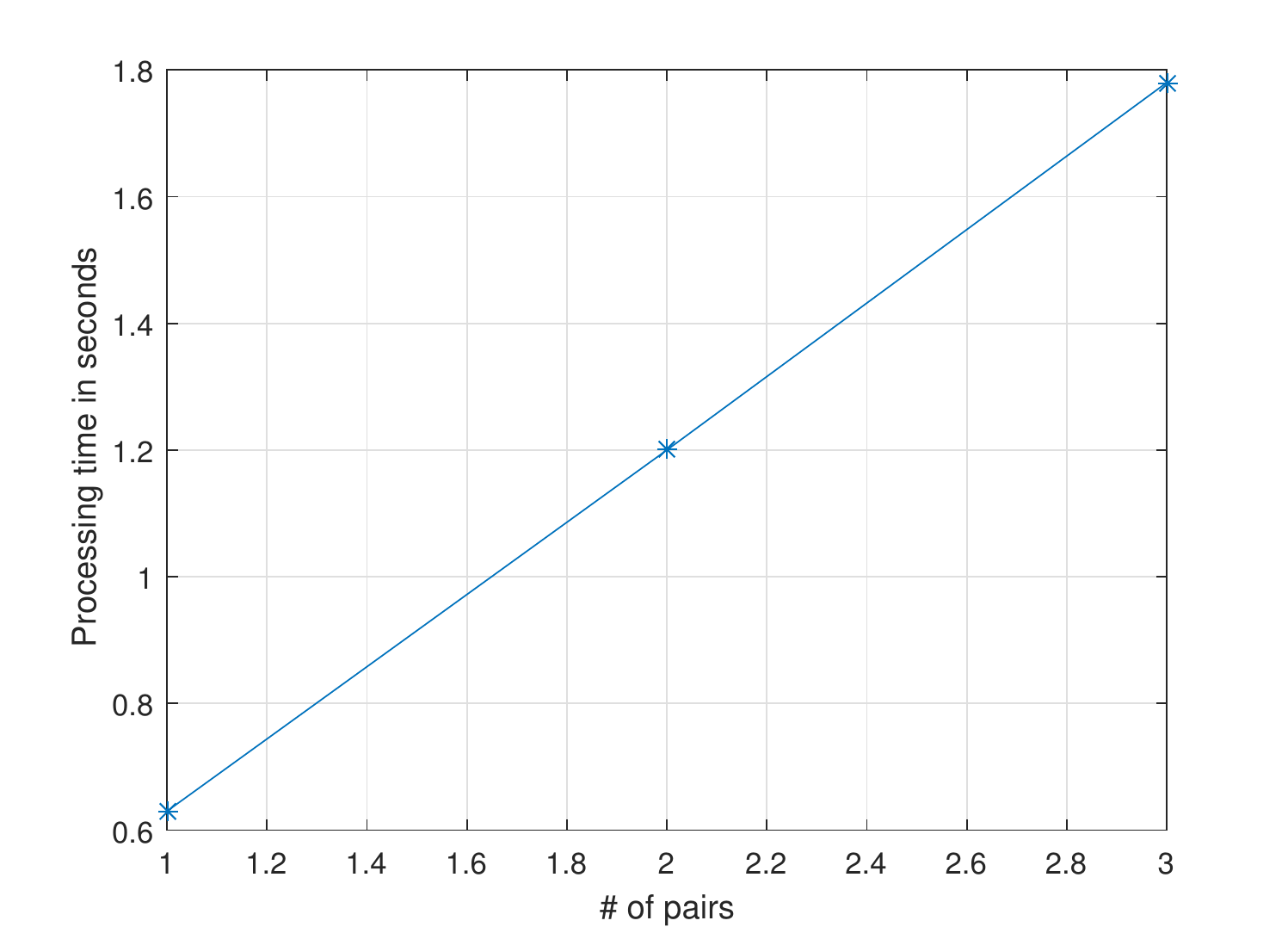}
		\caption{Processing Time with different number of pairs}
		\label{fig:sub3}
	\end{subfigure}
	
	\caption{Processing time per resolution, views and number of pairs.}
	\label{fig:test}
\end{figure}

\begin{table}[h!]
	\centering
	\begin{tabular}{||c c c ||} 
		\hline
		Pair & SSIM & PSNR \\ [0.5ex] 
		\hline\hline
		 (node 1, node 2) & 1.0 & $\approx infinity$  \\ 
		(node 1, node 3) & 1.0 & $\approx infinity$ \\
		(node 1, node 4) & 1.0 & $\approx infinity$  \\ 
		\hline 
	\end{tabular}
	\caption{SSIM and PSNR}
	\label{table:1}
\end{table}

The Strecha \textit{et al.}~\cite{strecha08} multi-view dataset has been used for comparison, while all simulations are computed using Matlab 2016. 
Figure~\ref{fig:Figure2} shows different views of the same scene.  Disparity maps using SAD and SSD are depicted in Figure~\ref{fig:Figure3}. Figure~\ref{fig:sub1} emphasizes that SAD is better than SSD when focusing on time performance, while the computation time decreases with the captured images resolutions. Note that the performance is independent from sensor coupling. This is obvious in Figure~\ref{fig:sub2}, where the processing time is approximately the same for different views. Finally, Figure~\ref{fig:sub3} shows how processing time increases linearly with the number of chosen couples. 

Table~\ref{table:1} contains structural similarity (SSIM) and Peak Signal-to-Noise Ratio (PSNR) functions applied on different pairs. SSIM and PSNR are used to measure the similarity between the two calculated disparity maps (using SAD or SSD). Getting 1.0 and infinity for different pairs ensure that the calculated disparity maps have approximately the same quality and are too similar. This will help us later to choose SAD as the best method -- because it has the same quality as SSD, but with a lower complexity leading to a more efficient processing.

Indeed, the complexity of the two methods is accessible theoretically, proving that SSD is more complex than SAD. In this first situation, WMSNs will need more time to process more complex algorithms, and then it will consume more energy~\cite{Mao09}. Computing the complexity of the two approaches using Equations~\ref{eq:3} and \ref{eq:4}, we have obtained the following results. For the SAD, we have 2 loops, one for horizontal width and the second one for vertical height of the image, while the operation within the loop is a single subtraction. So the SAD complexity is $O(n^2)$ in terms of elementary operations. The SSD, for its part, integrates a square within the same loop, increasing the complexity to $O(n^3)$ elementary operations, where $n$ is the number of lines (or columns) in the images. Such complexities are coherent with the simulations, leading to the choice of SAD as best compromise to achieve an acceptable depth evaluation in a WMSN-based video surveillance context.

\section{Conclusion and future works}
%WMSN is an encouraging field of research applied on different domains to resolve multiple needs. 
Disparity map is a main parameter to get the depth of a monitored scene and then reconstruct it.	In this research article, we contributed by applying disparity map calculation on WMSN distributed nodes. Our approach is directed by the main WMSN limitations: energy consumption, processing capability, QoS, and communication. The experiments and studies we done help to choose the best disparity map calculation method for WMSN usage, namely the SAD approach.

In future work, we intend to investigate not the existing literature, but novel disparity map computation methods specifically designed for wireless multimedia sensor networks. They will reach the optimized compromise between quality of the maps and complexity to obtain them. We will further investigate the optimal way to transfer the map from the terminal couple of nodes until the sink, by updating the disparities at aggregator nodes when receiving maps from various close couples of sensors (that observe a similar scene). The proposal will finally be distributed in a real wireless sensor networks, in order to test \textit{in vivo} the real performance of this optimized surveillance in operational context.


\balance
\section*{Acknowledgment}

\textit{This work is partially funded by the Labex ACTION program (contract ANR-11-LABX-01-01), the France-Suisse Interreg RESponSE project, and the National Council for Scientific Research in Lebanon.}



%\bibliographystyle{IEEEtran}

\bibliographystyle{IEEEtran}
\bibliography{references}

\end{document}